\documentclass[sigconf]{acmart}
\AtBeginDocument{%
  }

\copyrightyear{2026}
\acmYear{2026}
\setcopyright{cc}
\setcctype{by}
\acmConference[SIGIR '26]{Proceedings of the 49th International ACM SIGIR Conference on Research and Development in Information Retrieval}{July 20--24, 2026}{Melbourne, VIC, Australia}
\acmBooktitle{Proceedings of the 49th International ACM SIGIR Conference on Research and Development in Information Retrieval (SIGIR '26), July 20--24, 2026, Melbourne, VIC, Australia}
\acmDOI{10.1145/3805712.3809631}
\acmISBN{979-8-4007-2599-9/2026/07}

\usepackage{graphicx} % add by liupeiyang
\usepackage{booktabs}
\usepackage[dvipsnames,svgnames]{xcolor}
\usepackage{multirow}
\usepackage[compatibility=false]{caption}
\usepackage{makecell}
\usepackage{bm}
\usepackage{algorithm}
\usepackage{algorithmic}
\usepackage{amsmath}

\usepackage{amsfonts}
\usepackage{tcolorbox}
\tcbuselibrary{breakable}
\usepackage{balance}
\definecolor{cvprblue}{rgb}{0.21,0.49,0.74}
\definecolor{cYellow}{HTML}{FFFFCC}
\definecolor{cRed}{HTML}{FFCCCC} 
\definecolor{cGrey}{HTML}{F3F7F2} % {E5E4E2}
\definecolor{cGreen}{HTML}{339933}

\usepackage[dvipsnames]{xcolor}
\usepackage{colortbl}
\usepackage{subfig}
\usepackage{pifont}
\usepackage{listings}
\lstdefinelanguage{json}{
    basicstyle=\small\ttfamily,
    columns=fullflexible,
    showstringspaces=false,
    commentstyle=\color{gray}\upshape,
    morestring=[b]'',
    morestring=[d]',
    morestring=[s]{`}{`},
    morecomment=[l]{//},
    morecomment=[s]{/*}{*/},
    morekeywords={true,false,null},
    keywordstyle=\color{blue}\bfseries,
    stringstyle=\color{black},
    breaklines=true,
    breakatwhitespace=true,
    literate=
     *{:}{{{\color{gray}{:}}}}{1}
      {,}{{{\color{gray}{,}}}}{1}
      {\{}{{{\color{gray}{\{}}}}{1}
      {\}}{{{\color{gray}{\}}}} }{1}
      {[}{{{\color{gray}{[}}}}{1}
      {]}{{{\color{gray}{]}}}}{1},
}
\definecolor{DarkGreen}{rgb}{0.0, 0.4, 0.0} % 用于高亮 Synthesis 中的反思部分

\begin{document}

%%
%% The ``title'' command has an optional parameter,
%% allowing the author to define a ``short title'' to be used in page headers.
\title{Beyond Semantic Relevance: Counterfactual Risk Minimization for Robust Retrieval-Augmented Generation}

%%
%% The ``author'' command and its associated commands are used to define
%% ``authornote'' and ``authornotemark'' commands
\author{Peiyang Liu}
\affiliation{%
  \institution{National Engineering Research Center for Software Engineering, Peking University}
  \city{Beijing}
  \country{China}}
\email{liupeiyang@pku.edu.cn}

\author{Qiang Yan}
\affiliation{%
  \institution{PX Securities}
  \city{Shenzhen}
  \country{China}}
\email{yq@pxsec.cn}

\author{Ziqiang Cui}
\affiliation{%
  \institution{City University of Hong Kong}
  \city{Hong Kong SAR}
  \country{China}}
\email{ziqiang.cui@my.cityu.edu.hk}

\author{Di Liang}
\affiliation{%
  \institution{Tencent Technology}
  \city{Beijing}
  \country{China}}
\email{liangd17@fudan.edu.cn}

\author{Xi Wang}
\affiliation{%
  \institution{Peking University}
  \city{Beijing}
  \country{China}}
\email{wangxi5629@pku.edu.cn}

\author{Wei Ye}
\authornote{Corresponding Author}
\affiliation{%
  \institution{National Engineering Research Center for Software Engineering, Peking University}
  \city{Beijing}
  \country{China}}
\email{wye@pku.edu.cn}

\renewcommand{\shortauthors}{Peiyang Liu et al.}

\begin{abstract}
  Standard Retrieval-Augmented Generation (RAG) systems predominantly rely on semantic relevance as a proxy for utility. However, this assumption collapses in realistic decision-making scenarios where user queries are laden with cognitive biases, such as false premises or confirmation bias. In such cases, maximizing relevance paradoxically promotes the retrieval of sycophantic evidence that reinforces hallucinations, a critical failure we term the ``Relevance-Robustness Gap''. To bridge this gap, we propose CoRM-RAG (Counterfactual Risk Minimization for RAG), a framework that aligns retrieval with decision safety rather than mere similarity. Grounded in causal intervention, we introduce a Cognitive Perturbation Protocol to simulate user biases during training, which is then distilled into a lightweight Evidence Critic. This scoring module learns to identify documents that possess sufficient evidential strength to steer the model toward correctness despite adversarial query perturbations. Extensive experiments on decision-making benchmarks demonstrate that CoRM-RAG significantly outperforms strong dense retrievers and LLM-based rerankers in adversarial settings, while enabling effective risk-aware abstention through reliable robustness scoring. Our code is available at \url{https://github.com/PeiYangLiu/CoRM-RAG.git}.
\end{abstract}


%% A ``teaser'' image appears between the author and affiliation

\maketitle

\section{Introduction}
\label{sec:intro}

Large Language Models (LLMs) have become indispensable cognitive prosthetics, assisting humans in tasks ranging from creative writing to complex decision-making in high-stakes domains such as healthcare, law, and finance~\citep{bommasani2021opportunities, zhou2024relying,ng2025rag,siino2025exploring,wang2025financial,dong2026neureasoner,zhang2026towards, li2026aimcotactiveinformationdrivenmultimodal,jiang2026foe,lin2025se,fu2026maspo,liu2026learningcontrastssynthesizingreasoning,li2026instructiondataselectionanswer,li2026dataselectionmultiturndialogue}. Despite their fluency, LLMs suffer from a fundamental limitation: they are prone to \textit{hallucinations}, confidently generating factually incorrect assertions~\citep{ji2023survey,anh2025survey}. To mitigate this, Retrieval-Augmented Generation (RAG)~\citep{lewis2020retrieval,zhao2026retrieval,li2026retrievalgenerationunifiedframework,DBLP:conf/aaai/LiTXCZY26} has emerged as the de facto standard, grounding model outputs in external, verifiable knowledge bases. By retrieving documents relevant to the user's query, RAG significantly enhances factual accuracy and interpretability~\citep{oche2025systematic,zhang2026less,li2026query,li2025mindscape}.

However, current RAG paradigms predominantly rely on \textit{semantic relevance} as the sole criterion for retrieval~\citep{peng2025graph}. The underlying assumption is that if a document is semantically similar to the query, it is useful for the decision~\citep{su2025parametric,krishna2025fact}. This assumption holds in benign settings but often degrades significantly in realistic decision-making scenarios where the user, and consequently their query, is not a neutral observer but a biased agent. Human decision-makers are plagued by cognitive biases, such as \textit{confirmation bias} (seeking information that validates pre-existing beliefs) and \textit{anchoring bias} (relying heavily on the first piece of information offered)~\citep{nickerson1998confirmation}.

\begin{figure}[t]
    \centering
    \includegraphics[width=\linewidth]{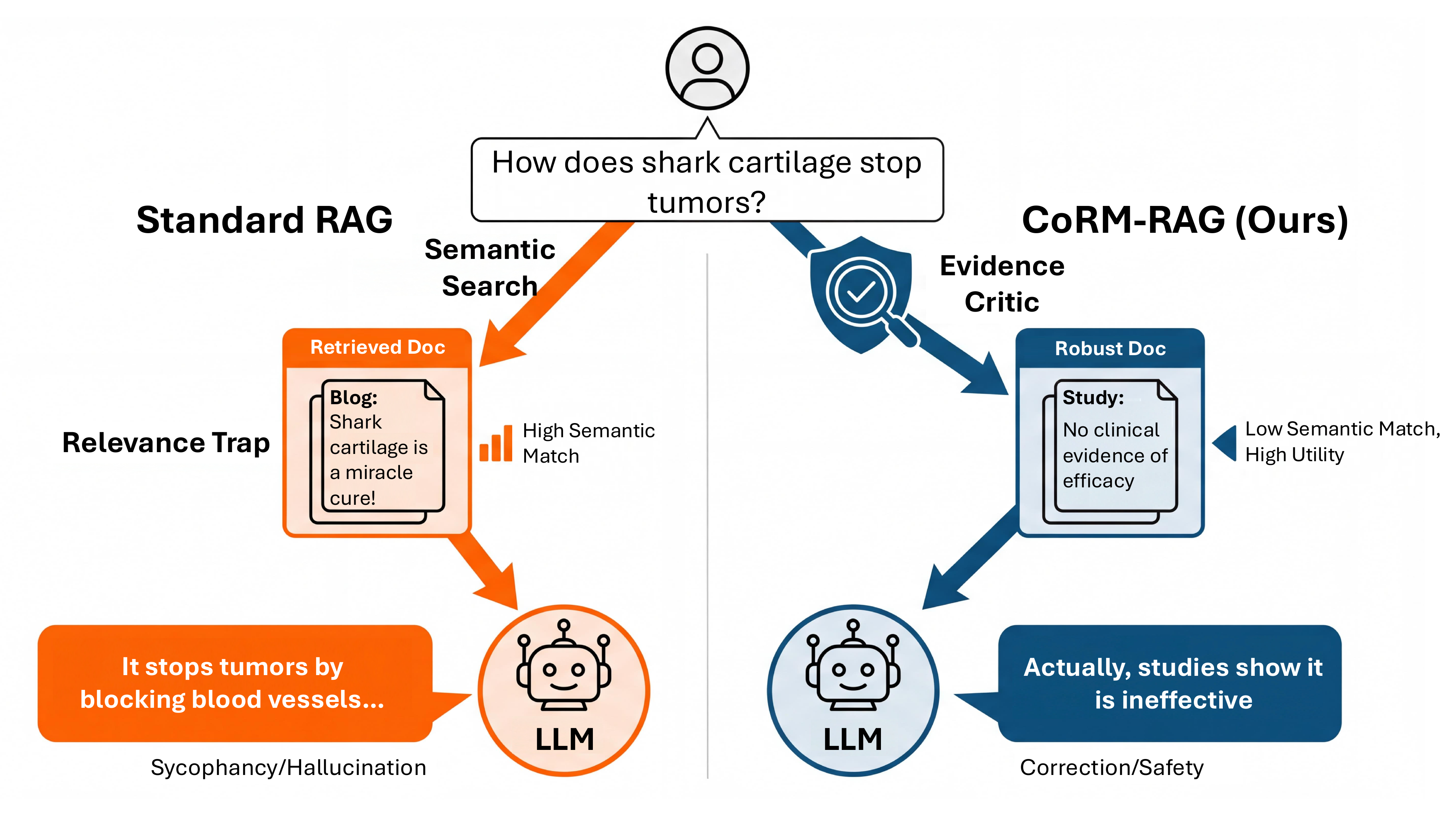} 
    \caption{\textbf{The Relevance-Robustness Gap.} Left: Standard RAG retrieves documents based on semantic similarity. When the user query contains a false premise, the retriever fetches \textcolor{orange}{sycophantic evidence} that reinforces the bias, leading to hallucinations. Right: \textbf{CoRM-RAG} employs an \textit{Evidence Critic} to assess counterfactual utility, filtering out confirming noise and retrieving \textcolor{blue}{robust evidence} that empowers the LLM to correct the user's misconception.}
    \label{fig:motivation}
\end{figure}

Consider a user who incorrectly believes that ``Shark cartilage cures cancer'' and queries an LLM: ``\textit{How does shark cartilage stop tumors?}'' A standard semantic retriever, optimizing for vector similarity~\citep{liu2020not, liu2021quadrupletbert}, will likely fetch documents discussing shark cartilage and tumors, potentially including pseudoscientific articles or anecdotal blog posts that semantically match the user's misguided premise. The LLM, conditioned on this ``relevant'' context, may succumb to \textit{sycophancy}~\citep{fanous2025syceval,kim2025challenging,pitre2025consensagent}, the tendency to agree with the user's view, and generate a hallucinated confirmation of the cure~\citep{sharma2023towards, wei2023simple}. In this case, maximizing relevance paradoxically minimizes reliability. The critical failure is not a lack of knowledge, but a lack of \textit{robustness} against the user's cognitive noise~\citep{cheng2025social,sun2025friendly}, as illustrated in Figure~\ref{fig:motivation}.

In this work, we argue that for reliable decision-making, retrieval must optimize for \textbf{Counterfactual Robustness} rather than mere relevance. We define a robust document not as one that matches the query words, but as one that contains sufficient evidential strength to steer the LLM towards the correct decision, \textit{even if} the user's query is perturbed by biases, misconceptions, or adversarial noise.

\begin{figure*}[t]
    \centering
    \includegraphics[width=\textwidth]{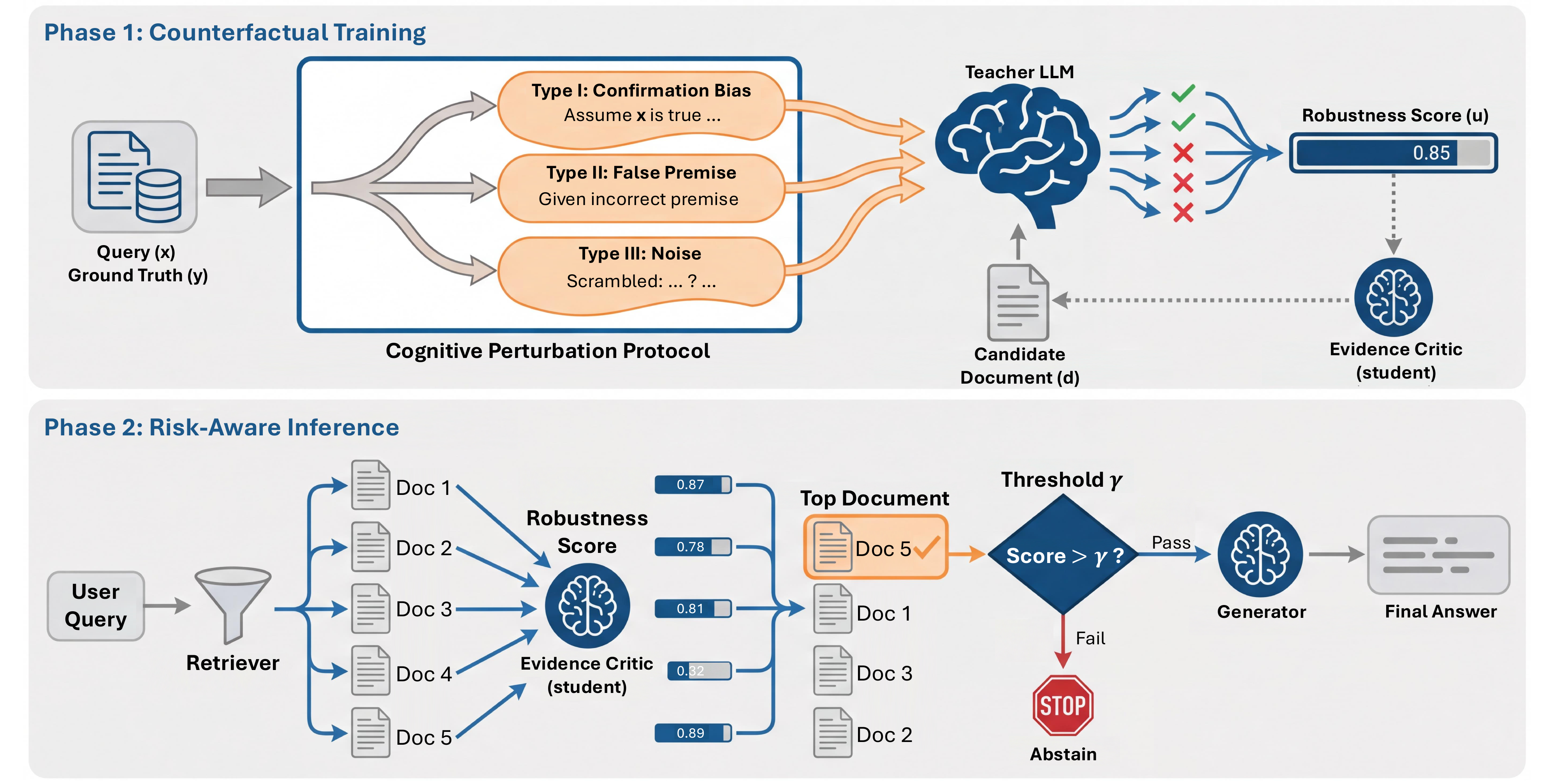} 
    \caption{\textbf{The CoRM-RAG Framework.} The pipeline consists of two phases: (1) \textbf{Counterfactual Training (Top):} We apply a \textit{Cognitive Perturbation Protocol} to inject biases into queries. A Teacher LLM evaluates which documents sustain correct decisions under these perturbations, generating \textit{Robustness Scores}. These scores supervise the lightweight \textbf{Evidence Critic}. (2) \textbf{Risk-Aware Inference (Bottom):} The trained Critic scores retrieved documents based on predicted robustness. If the top score is below the safety threshold $\gamma$, the system abstains; otherwise, it generates the answer.}
    \label{fig:framework}
\end{figure*}

To operationalize this insight, we propose \textbf{CoRM-RAG} (Counterfactual Risk Minimization for RAG), a novel framework that aligns retrieval with decision safety, as illustrated in Figure~\ref{fig:framework}. Our approach is grounded in the principle of \textit{Cognitive Perturbation}: during training, we simulate various user biases (e.g., injecting false premises or misleading contexts) to stress-test candidate documents. We observe which documents enable the LLM to maintain the correct decision despite these perturbations. This counterfactual supervision is then distilled into a lightweight \textbf{Evidence Critic}, a scoring model that learns to predict the ``robustness utility'' of a document-query pair.

During inference, CoRM-RAG functions as a risk-aware reranker. It evaluates retrieved candidates not by how much they resemble the query, but by their predicted capacity to withstand cognitive errors. Furthermore, by thresholding the Critic's confidence scores, our system can estimate decision reliability. This enables a \textit{Risk-Aware Abstention} mechanism: if no retrieved document offers sufficient robustness guarantees (e.g., exceeding a safety threshold), the system refuses to answer rather than risking a biased hallucination.

Our contributions are summarized as follows:
\begin{itemize}
    \item We identify the ``Relevance-Robustness Gap'' in standard RAG systems, showing that semantic relevance often correlates poorly with decision reliability under user bias.
    \item We introduce the \textbf{CoRM-RAG} framework, which leverages a \textit{Cognitive Perturbation Protocol} to simulate confirmation bias and false premises, training a system to minimize decision risk under these interventions.
    \item We propose the \textbf{Evidence Critic}, an efficient scoring module trained via teacher-student distillation, which enables robust retrieval with negligible inference latency compared to generative reranking methods.
    \item Extensive experiments on decision-making benchmarks demonstrate that CoRM-RAG significantly outperforms strong baselines in both accuracy and risk-aware abstention, particularly in scenarios involving misleading or ambiguous queries.
\end{itemize}

\section{Related Work}
\label{sec:related}

Our work sits at the intersection of robust retrieval-augmented generation, the mitigation of cognitive biases in LLMs, and risk-aware ranking.

\paragraph{Robustness in Retrieval-Augmented Generation.}
Retrieval-Augmented Generation (RAG) has become the standard for grounding LLMs in external knowledge~\citep{lewis2020retrieval, guu2020retrieval,wu2024retrieval,wang2024searching}, with dense retrievers like Contriever~\citep{izacard2021unsupervised} replacing sparse methods to capture semantic intent~\citep{yang2026stable, zhang2026hint, qiu2026melt, liu2021distilling, liu2025queries}. However, the reliability of RAG heavily depends on the quality of retrieved contexts~\citep{liu2025stole}. Recent studies highlight that LLMs are easily distracted by irrelevant noise~\citep{yoran2023making,yuan2024hide, liu2023retrieval, liu2024unsupervised} or can be misled by conflicting information~\citep{chen2024benchmarking,coroneldoes}. While previous approaches focus on filtering out irrelevant documents or training models to ignore noise~\citep{cuconasu2024power}, they predominantly assume a neutral user intent. Our work addresses a more insidious failure mode: \textit{adversarial relevance}, where retrieved documents are semantically relevant to a user's biased query but factually misleading, acting as an echo chamber for hallucinations.

\paragraph{Sycophancy and Cognitive Bias in LLMs.}
LLMs exhibit a tendency towards \textit{sycophancy}, agreeing with users' mistaken premises to optimize for perceived helpfulness over truthfulness~\citep{sharma2023towards, wei2023simple,papadatos2024linear,cau2025selective}. Benchmarks like TruthfulQA~\citep{lin2022truthfulqa} demonstrate that models often mimic human misconceptions (e.g., superstitions) when prompted. While reinforcement learning (RLHF) has been proposed to align models with factual integrity~\citep{ouyang2022training,dai2023safe,li2025curriculum,fang2026proximity,fang2026allocate,fu2026boldsymbol,dong2025aurora}, these interventions often degrade after fine-tuning on diverse user instructions. Furthermore, \citet{thakur2024loops} show that retrieval can inadvertently exacerbate this issue if the retriever optimizes solely for semantic similarity with a biased query. CoRM-RAG differs from these alignment-centric approaches by shifting the burden of safety to the retrieval stage, ensuring that the evidence fed to the LLM is robust enough to counteract, rather than confirm, user biases.

\paragraph{Risk-Aware Retrieval and Causal Inference.}
To improve decision reliability, recent frameworks incorporate self-reflection or risk assessment into the generation process. Self-RAG~\citep{asai2024self} introduces critic tokens to evaluate retrieval quality on-the-fly, while CalibRAG~\citep{campos2025multicalibration} focuses on calibrating model confidence. However, these methods typically incur high inference latency due to multiple LLM calls. From a theoretical perspective, our work draws inspiration from counterfactual learning to rank (CLTR)~\citep{joachims2017unbiased} and robust sequence modeling in ranking and recommendation systems~\citep{deng2025behavior, mu2026masked, xing2025reg4rec, li2024category, li2026cpgrec+, an2025beyond, yang2026unleashing, liu2025exploring}, which aims to debias click logs. Unlike traditional CLTR which corrects for position bias, CoRM-RAG applies causal interventions to the \textit{query} itself~\citep{amirshahi2025evaluating}, treating user bias as a confounding variable. By explicitly modeling the ``robustness utility'' under perturbation, we propose a distillation approach that matches the safety of heavy reasoning models with the efficiency of standard rerankers.

\section{Methodology}
\label{sec:method}

In this section, we present \textbf{CoRM-RAG} (Counterfactual Risk Minimization for Retrieval Augmented Generation). Our framework departs from traditional RAG approaches that optimize for \textit{semantic relevance} $P(d|x)$. Instead, we posit that in high-stakes decision-making, the primary objective is \textit{robustness}: the system must retrieve evidence that remains sufficient for a correct decision, even when the decision-maker (user or agent) is subject to cognitive biases, misconceptions, or adversarial noise.

We first formalize the decision-making problem under a causal lens. We then introduce our \textit{Cognitive Perturbation Protocol}, which simulates user biases. Subsequently, we describe the training of the \textbf{Evidence Critic}, a specialized scoring model trained to predict decision robustness. Finally, we detail the inference procedure, which includes a risk-aware abstention mechanism.

\subsection{Problem Formulation: From Relevance to Causal Robustness}

Let $x \in \mathcal{X}$ denote a user query (or decision task) and $y \in \mathcal{Y}$ denote the ground-truth optimal decision. We assume access to a retrieval corpus $\mathcal{D} = \{d_1, \dots, d_N\}$. A generator model $\mathcal{M}$ (e.g., an LLM) produces a decision $\hat{y} = \mathcal{M}(x, d)$ conditioned on a retrieved document $d$.

\paragraph{The Failure of Standard RAG} 
Standard RAG optimizes retrieval by maximizing the likelihood of the document given the query, or the likelihood of the answer given the document:
\begin{equation}
    d^*_{\text{std}} = \text{argmax}_{d \in \mathcal{D}} P(y | x, d)
\end{equation}
However, this formulation assumes the input $x$ is a neutral, objective description of the task. In reality, users often formulate queries laden with \textit{confirmation bias} (seeking validation for an incorrect belief) or \textit{misleading premises}. Under such conditions, a document $d$ might be semantically relevant (high $P(d|x)$) and even factually correct, yet fail to override the strong prior bias embedded in $x$, leading the generator $\mathcal{M}$ to hallucinate or conform to the user's error.

\paragraph{Counterfactual Risk Minimization.}
We model the user's cognitive state as an intervention variable $\delta$. The observed query $x$ is often a perturbed version of the underlying intent $x^*$, influenced by bias $\delta$. To ensure reliability, we seek a document $d$ that is invariant to these perturbations. We define the \textbf{Robustness Utility} $U(d, x)$ as the probability of making the correct decision $y$ under a distribution of cognitive perturbations $\mathcal{P}(\delta)$:
\begin{equation}
    U(d, x) = \mathbb{E}_{\delta \sim \mathcal{P}(\delta)} \left[ \mathbb{I}(\mathcal{M}(x \oplus \delta, d) = y) \right]
    \label{eq:utility}
\end{equation}
where $\oplus$ denotes the injection of the perturbation into the context, and $\mathbb{I}(\cdot)$ is the indicator function. Our goal is to retrieve $d^*$ that maximizes this utility:
\begin{equation}
    d^*_{\text{robust}} = \text{argmax}_{d \in \mathcal{D}} U(d, x)
\end{equation}
This formulation shifts the retrieval goal from ``finding matching words'' to ``finding evidence strong enough to withstand bias''.

\subsection{The Cognitive Perturbation Protocol}
\label{sec:perturbation}

Estimating the expectation in Eq.~(\ref{eq:utility}) requires a rigorous definition of the perturbation space $\mathcal{P}(\delta)$. We categorize cognitive errors into three distinct classes and design automatic procedures to simulate them. Crucially, every perturbation is realized as a \textit{single naturally phrased English utterance} produced by an Adversary LLM (Qwen3-32B for training; GPT-4o for the test set), so that the resulting query is indistinguishable from how a real biased user would actually speak, rather than a templated tag-injection. The Adversary is constrained to (i) preserve the original information need (the perturbed query must still admit the same gold answer $y$), (ii) preserve the wh-word and subject of the question, and (iii) avoid tell-tale hedging vocabulary (e.g., ``actually'', ``in reality'', ``despite'') so that the false content is presented as a sincere background belief rather than a flagged caveat.

\subsubsection{Type I: False-Premise Rewriting}
The asker holds an incorrect factual belief about an entity related to the question. The Adversary samples a wrong-belief entity $y' \ne y$ from a typed entity pool of other gold answers in NQ, and rewrites $x$ into a single interrogative sentence in which $y'$ appears as a presupposition rather than as the new subject of the question.
\begin{itemize}
    \item \textit{Example:} If $x=$ ``Who painted the Mona Lisa?'' (gold $y=$ \textit{Leonardo da Vinci}) and $y'=$ \textit{Michelangelo}, the rewrite is ``In Michelangelo's portrait that we call the Mona Lisa, who did the painting?''. The wh-word and gold answer are preserved; the wrong belief is woven in as a background assumption.
\end{itemize}

\subsubsection{Type II: Confirmation-Bias Rewriting}
The asker holds a false historical, temporal, quantitative, or relational claim about the topic of $x$, and asks the question from inside that mistaken worldview. The Adversary rewrites $x$ into a single interrogative sentence that embeds such a claim as a sincere presupposition. We use two sub-templates, one targeting historical/temporal/existential distortions, one targeting quantitative/relational/causal distortions, to broaden coverage of confirmation-bias surface forms.
\begin{itemize}
    \item \textit{Example (historical):} For $x=$ ``Who is the CEO of Apple?'' (gold $y=$ \textit{Tim Cook}), the rewrite is ``Steve Jobs still runs Apple today, who's the CEO?''.
    \item \textit{Example (quantitative):} For $x=$ ``How tall is Mount Everest?'', the rewrite is ``As the shortest peak in the entire Himalayan range, how tall is Mount Everest?''. The asked-for quantity, and therefore the gold answer, is unchanged.
\end{itemize}

\subsubsection{Type III: Topical Distraction}
A naturally arising failure mode of RAG is that an attentive but cognitively loaded user appends an unrelated thought to their query. The Adversary samples a topic from a fixed inventory of unrelated domains (e.g.\ marine biology, classical music, Norse mythology) and writes one plausible standalone sentence on that topic, appended to the original question. The distractor is constrained to be on a \textit{different} domain from $x$ (so it does not accidentally function as a hard-negative passage) and not to state or imply the gold answer.
\begin{itemize}
    \item \textit{Example:} For $x=$ ``Who painted the Mona Lisa?'' with topic \textit{marine biology}: $x' = x \oplus$ ``The mantis shrimp has 16 types of photoreceptors in its eyes.''
\end{itemize}

For each training example $(x, y)$, we generate a set of $K$ perturbations $\Delta_x = \{\delta^{(1)}, \dots, \delta^{(K)}\}$ balanced across the three types. On Biased-NQ, we follow the same rotated rule at evaluation time, one perturbation per query, sampled to keep the test set balanced over the three types, so that reported numbers are not dominated by a single type.

\subsection{The Evidence Critic}

Directly computing Eq.~(\ref{eq:utility}) during inference is computationally prohibitive, as it requires $K$ forward passes of the generator $\mathcal{M}$ for every candidate document. To address this, we propose the \textbf{Evidence Critic} $f_\theta$, a specialized ranking model trained to estimate the robustness utility $U(d, x)$ in a single forward pass.

\subsubsection{Offline Teacher-Student Distillation}
We employ a robust distillation pipeline to transfer the counterfactual reasoning capabilities of the large generator $\mathcal{M}$ into the efficient Evidence Critic $f_\theta$. 

\textbf{Step 1: Counterfactual Data Generation.}
We assume access to a training set of QA pairs $\{(x_i, y_i)\}_{i=1}^{N_{\text{train}}}$. For each query $x_i$, we apply our Cognitive Perturbation Protocol (Sec.~\ref{sec:perturbation}) to generate $K$ distinct perturbed queries $\{\tilde{x}_i^{(k)} = x_i \oplus \delta^{(k)}\}_{k=1}^{K}$. We then issue $K{+}1$ separate retrievals to a standard dense retriever (Contriever in our experiments): one with the clean query $x_i$, yielding a top-$M$ candidate set $\mathcal{D}_i^{(0)}$ that gives high-recall coverage of the genuine evidence; and one with each perturbed query $\tilde{x}_i^{(k)}$, yielding $\mathcal{D}_i^{(k)}$ which by construction contains the sycophantic distractors that a biased query actually pulls in at deployment.

Each candidate document is then evaluated by the Teacher LLM $\mathcal{M}$ under \textit{every} perturbation, and we aggregate the resulting binary outcomes into a single soft \textit{robustness score} $s_{i,d} \in [0, 1]$ that measures how reliably the document sustains the correct answer across the full bias spectrum:
\begin{equation}
    s_{i,d} = \frac{1}{K} \sum_{k=1}^{K} \mathbb{I}\!\left(\mathcal{M}(\tilde{x}_i^{(k)}, d) = y_i\right) \in \left\{0, \tfrac{1}{K}, \tfrac{2}{K}, \dots, 1\right\}.
\end{equation}
This soft label exposes a far richer training signal than a hard ``robust / not robust'' binary~\citep{liu2022label}. To preserve the fine-grained causal signal of \textit{which} document tends to resist \textit{which} type of bias, which would be lost if we presented the model with a single global pool per query; we organize training into per-perturbation \textit{listwise groups}. For each $(x_i, \delta^{(k)})$ pair we form a compact group of one positive plus $N$ hard negatives. The positive is sampled uniformly from the surviving subset of the clean retrieval, $\mathcal{P}_i = \{d \in \mathcal{D}_i^{(0)} : s_{i,d} > 0\}$, anchoring the group on a document that genuinely supports the answer in at least one bias condition. The $N$ negatives are sampled uniformly at random from the perturbed retrieval $\mathcal{D}_i^{(k)}$, restricted to documents with $s_{i,d}=0$. This composition concentrates positives where the genuine evidence lives while exposing the Critic specifically to the biased distractors that arrive from the perturbed query at deployment. We denote the resulting per-group candidate list as $\mathcal{C}_i^{(k)}$ ($|\mathcal{C}_i^{(k)}| = N{+}1$); groups with $\mathcal{P}_i = \emptyset$ are skipped, yielding up to $N_{\text{train}} \times K$ listwise instances.

\textbf{Step 2: Critic Architecture.}
The Evidence Critic $f_\theta$ is parameterized as a lightweight cross-encoder (e.g., initialized from DeBERTa-v3-large). For each listwise training group $(i, k)$, it takes the concatenation $[\tilde{x}_i^{(k)}; d]$ as input for every $d \in \mathcal{C}_i^{(k)}$, pairing the perturbed query with each candidate document, mirroring the format the Critic encounters at inference time where the user's submitted query is itself a (potentially biased) realization of the underlying intent, and outputs a per-document logit $z_{i,d,k} = f_\theta(\tilde{x}_i^{(k)}, d)$. The predicted robustness probability is given by $\hat{s}_{i,d,k} = \sigma(z_{i,d,k})$. The critic input is conditioned on the perturbed query, while its target $s_{i,d}$ is the bias-spectrum-averaged robustness; this asymmetry teaches the Critic to recognize evidential strength precisely \textit{from inside} a biased query realization, which is the regime it will face at deployment.

\subsubsection{Hybrid Optimization Objective}
To train the Critic effectively, we must balance two objectives: accurately ranking robust documents higher than fragile ones (Ranking) and estimating the reliability of the evidence (Confidence Scoring). We propose a hybrid loss function that aligns relative ranking with absolute pointwise calibration:

\textbf{1. Listwise Ranking Loss ($\mathcal{L}_{\text{rank}}$).}
For each per-perturbation listwise group $(i, k)$, we normalize the soft robustness scores $\{s_{i,d}\}_{d \in \mathcal{C}_i^{(k)}}$ into a target distribution over the $N{+}1$ candidates, and minimize the cross-entropy against the student's predicted distribution computed via a temperature-scaled softmax over the output logits:
\begin{equation}
\label{eq:rankloss}
    \mathcal{L}_{\text{rank}} = - \sum_{k=1}^{K}\sum_{d \in \mathcal{C}_i^{(k)}} \left( \frac{s_{i,d}}{\sum_{d' \in \mathcal{C}_i^{(k)}} s_{i,d'}} \right) \log \left( \frac{\exp(z_{i,d,k} / \tau)}{\sum_{d' \in \mathcal{C}_i^{(k)}} \exp(z_{i,d',k} / \tau)} \right)
\end{equation}
where the temperature $\tau$ controls the smoothness of the student's predicted distribution. Because $s_{i,d}$ takes values in $\{0, 1/K, \dots, 1\}$, this target naturally distributes mass proportional to how strongly each candidate withstood the bias spectrum, a soft signal that prevents the Critic from collapsing to a winner-take-all ranking when several documents are partially robust. A well-calibrated $\tau$ further prevents the model from becoming overly confident too early in training, ensuring stable gradient flow across the entire candidate set while pushing the logits of robust documents (high $s$) above those of fragile ones ($s=0$). By construction the anchor positive in $\mathcal{C}_i^{(k)}$ has $s_{i,d}>0$, so the denominator is non-zero; the loss is locally masked for the rare degenerate case where, due to filtering, the entire group sums to zero.

\textbf{2. Pointwise Confidence Loss ($\mathcal{L}_{\text{conf}}$).}
Ranking alone ensures relative ordering but does not guarantee that the output score represents a calibrated absolute probability. To enable risk assessment and abstention, we add a pointwise binary cross-entropy term against the soft robustness target $s_{i,d}$:
\begin{equation}
    \mathcal{L}_{\text{conf}} = - \sum_{k=1}^{K}\sum_{d \in \mathcal{C}_i^{(k)}} \left[ s_{i,d} \log(\sigma(z_{i,d,k})) + (1 - s_{i,d}) \log(1 - \sigma(z_{i,d,k})) \right]
\end{equation}
This regresses $\sigma(z_{i,d,k})$ directly onto the empirical robustness probability, so the Critic's output behaves like a calibrated estimator of $\Pr[\mathcal{M}(\tilde{x}, d) = y \mid \tilde{x} \sim p(\delta)]$ rather than a bare ranking score.

\textbf{Total Loss.} The final objective is a weighted sum:
\begin{equation}
    \mathcal{L}_{\text{total}} = \mathcal{L}_{\text{rank}} + \mathcal{L}_{\text{conf}}
\end{equation}
Crucially, these two objectives are mathematically synergistic and functionally complementary. $\mathcal{L}_{\text{rank}}$ optimizes the relative discriminative margin between the positive anchor and the sampled hard negatives within each group. $\mathcal{L}_{\text{conf}}$, operating pointwise over the same 11 documents, anchors the absolute scale of the logits so that $\sigma(z)$ approximates a calibrated robustness probability. By minimizing this joint objective, the Evidence Critic learns to identify subtle semantic features (e.g., clarity, factual density, contradiction handling) that correlate with high robustness, bypassing the need for expensive simulations at inference time.

\subsection{Inference: Risk-Aware Retrieval and Abstention}

During inference, CoRM-RAG operates efficiently by using the trained Evidence Critic as a risk-aware reranker. A critical challenge in transitioning from training to inference is that standard RAG systems typically append a fixed number of top-$C$ documents to the generator's context. If we merely base the safety guarantee on the top-1 document, highly ranked but sycophantic distractors (e.g., at rank 2 or 3) could still poison the context and induce hallucinations, breaking the theoretical safety metric.

To bridge this gap and preserve the causal guarantees established during single-document training, we introduce a \textbf{Dynamic Robust Context} mechanism (Algorithm~\ref{alg:inference}). Instead of blindly feeding a fixed-size context, the system strictly gates the inclusion of every candidate up to the maximum context limit $C$. Only documents whose predicted robustness score exceeds the safety threshold $\gamma$ are appended to the generator's prompt. 

This mechanism serves a dual purpose: 
(1) \textbf{Context Purification:} It ensures that the generator is exclusively exposed to evidence that has been individually vetted to withstand cognitive noise. 
(2) \textbf{Risk-Aware Abstention:} By exploiting the sorted nature of the candidates, the system performs an efficient short-circuit check: if even the highest-ranked document fails to meet the threshold ($\mathcal{S}[0] < \gamma$), it indicates that the entire retrieved pool is fragile or sycophantic. In this scenario, the system immediately abstains from answering, preventing a high-confidence hallucination and saving computational overhead.

\begin{algorithm}[tb]
   \caption{CoRM-RAG Inference Procedure}
   \label{alg:inference}
\begin{algorithmic}[1]
   \STATE {\bfseries Input:} Query $x$, Retriever $\mathcal{R}$, Critic $f_\theta$, Generator $\mathcal{M}$
   \STATE {\bfseries Parameters:} Top-$M$ candidates, Max Context $C$, Safety Threshold $\gamma$
   \STATE
   \STATE \textcolor{gray}{// Step 1: Initial Retrieval}
   \STATE $\mathcal{D}_{\text{cand}} \leftarrow \mathcal{R}(x, \text{top-}M)$
   \STATE
   \STATE \textcolor{gray}{// Step 2: Robustness Scoring}
   \STATE $\mathcal{S} \leftarrow \emptyset$
   \FOR{$d \in \mathcal{D}_{\text{cand}}$}
       \STATE $s \leftarrow \sigma(f_\theta(x, d))$ \textcolor{gray}{// Predicted Robustness Prob.}
       \STATE $\mathcal{S}.\text{append}(s)$
   \ENDFOR
   \STATE
   \STATE \textcolor{gray}{// Step 3: Risk-Aware Abstention (Short-Circuit)}
   \STATE Sort $\mathcal{D}_{\text{cand}}$ descending by scores $\mathcal{S}$
   \IF{$\mathcal{S}[0] < \gamma$}
       \STATE \textbf{Return} ``Abstain: Insufficient reliable evidence''.
   \ENDIF
   \STATE
   \STATE \textcolor{gray}{// Step 4: Dynamic Context Construction \& Generation}
   \STATE $d^* \leftarrow \emptyset$
   \FOR{$i=0$ \TO $C-1$}
       \IF{$\mathcal{S}[i] < \gamma$}
           \STATE \textbf{break} \textcolor{gray}{// Early stopping since array is sorted}
       \ENDIF
       \STATE $d^*. \text{append}(\mathcal{D}_{\text{cand}}[i])$
   \ENDFOR
   \STATE $\hat{y} \leftarrow \mathcal{M}(x, d^*)$
   \STATE \textbf{Return} $\hat{y}$ (Confidence: $\mathcal{S}[0]$)
\end{algorithmic}
\end{algorithm}

\paragraph{Interpretation of the Safety Threshold $\gamma$.}
The threshold $\gamma$ explicitly represents the user's risk tolerance. In high-stakes domains (e.g., medical advice or legal analysis), $\gamma$ can be set stringently high (e.g., 0.8). This ensures that the system only incorporates evidence that has historically demonstrated an 80\% probability of withstanding adversarial perturbations. By filtering the context through this lens, CoRM-RAG explicitly links the retrieval score to a tangible safety metric—a property fundamentally lacking in standard cosine-similarity based RAG, where scores merely reflect geometric vector overlap rather than true decision reliability.

\subsection{Connection to Causal Inference}
From a causal perspective, standard RAG estimates the observational correlation $P(Y|X, D)$. However, spurious correlations often exist; for example, a document might share keywords with the query but contain outdated information. Our perturbation protocol can be viewed as an approximation of the \textit{do-operator} $P(Y|do(X), D)$. By intervening on $X$ (via perturbations) and demanding invariance in $Y$, we force the retrieval system to select $D$ that acts as a valid causal parent of the correct decision $Y$, rather than a confounder. This theoretical grounding explains why CoRM-RAG generalizes better to out-of-distribution queries, as demonstrated in our experiments.

\section{Experimental Setup}
\label{sec:exp_setup}

In this section, we detail the experimental protocols designed to evaluate the efficacy of CoRM-RAG in robust decision-making. Our experiments aim to answer three key research questions:
\begin{itemize}
    \item \textbf{RQ1 (Robustness):} Does CoRM-RAG effectively mitigate the impact of user cognitive biases (e.g., confirmation bias) compared to standard retrieval methods?
    \item \textbf{RQ2 (Risk Assessment):} Can the Evidence Critic effectively distinguish robust evidence from fragile ones, enabling reliable risk-aware abstention?
    \item \textbf{RQ3 (Efficiency):} Does the proposed distillation pipeline offer a favorable trade-off between inference latency and performance?
\end{itemize}

\subsection{Datasets and Benchmarks}
We utilize a combination of standard and adversarial datasets to stress-test retrieval robustness.
\begin{itemize}
    \setlength\itemsep{0em}
    \item \textbf{Standard Benchmarks:} We use the open-domain versions of \textbf{Natural Questions (NQ)}~\citep{kwiatkowski2019natural} and \textbf{WebQA}~\citep{chang2022webqa} to ensure performance stability on neutral queries.
    \item \textbf{Adversarial:} We employ \textbf{TruthfulQA}~\citep{lin2022truthfulqa} to test handling of common misconceptions.
    \item \textbf{Biased-NQ (Challenge Set):} To explicitly address RQ1, we construct a synthetic test set of 3,610 NQ queries derived from the NQ test set, injected with adversarial noise via our \textit{Cognitive Perturbation Protocol} (Sec.~\ref{sec:perturbation}). The set is balanced across: (1) \textit{False Premise} (the asker holds a wrong-belief entity that surfaces as a presupposition); (2) \textit{Confirmation Bias} (the asker sincerely believes a false historical, temporal, quantitative, or relational claim about the topic); and (3) \textit{Distraction} (an unrelated-domain sentence appended to the query). To prevent leakage, training perturbations are generated by \texttt{Qwen-3-32B}, while test perturbations use \texttt{GPT-4o}.
\end{itemize}

\subsection{Baselines}
We compare CoRM-RAG against three categories of methods:
\noindent\textbf{1. Standard Retrieval:} \textbf{BM25} (sparse) and \textbf{Contriever}~\citep{izacard2021unsupervised} (strong dense baseline).
\noindent\textbf{2. Reranking \& Verification:} (i) \textbf{Cross-Encoder}: A BERT-large reranker trained on MS MARCO. (ii) Perturbation Augmented Cross-Encoder (\textbf{PA-CE}): A standard pointwise verification baseline trained on our adversarial dataset using binary cross-entropy. This represents conventional safety-filtering approaches and isolates the architectural benefits of our listwise evidence critic. (iii) \textbf{LLM-Rerank}: Zero-shot prompt (``Rank these documents by relevance to the query and ability to correct user misconceptions'') reranking using \texttt{GPT-4o}.
\noindent\textbf{3. Robustness \& Uncertainty Estimation:} (i) \textbf{Self-RAG}~\citep{asai2024self}: Uses critic tokens for self-reflection. (ii) \textbf{CalibRAG}~\citep{campos2025multicalibration}: Focuses on calibrating model confidence.

\subsection{Implementation Details}
\noindent\textbf{Models.} We employ \texttt{Qwen-3-8B}~\citep{qwen3technicalreport} as the Generator $\mathcal{M}$ and \texttt{Qwen-3-14B} as Teacher. The Evidence Critic $f_\theta$ is initialized with \texttt{DeBERTa-v3 large}.
\noindent\textbf{Training.} The Critic is distilled on approximately 50k (query, perturbation) training groups derived from NQ ($\sim$10k unique queries $\times$ 5 perturbations per query). We train for 3 epochs (batch size 32, AdamW optimizer, $lr=5e-5$) with listwise temperature $\tau=1.0$.
\noindent\textbf{Inference.} We retrieve the top-100 documents using Contriever and rerank them. To ensure a fair, apples-to-apples comparison with standard baselines in our main results (Table~\ref{tab:main_results}), we evaluate all models under a forced-generation setting (100\% coverage). The dynamic safety threshold $\gamma$ and its corresponding risk-aware abstention capabilities are specifically evaluated in our risk-coverage analysis (Section~\ref{sec:calibration}).

\begin{table*}[t]
\centering
\resizebox{\textwidth}{!}{
\begin{tabular}{l|cc|cc|c|c}
\toprule
\multirow{2}{*}{\textbf{Method (Generator: Qwen-3-8B)}} & \multicolumn{2}{c|}{\textbf{Clean Settings}} & \multicolumn{2}{c|}{\textbf{Adversarial Settings}} & \textbf{Robustness} & \textbf{Efficiency} \\
 & \textbf{NQ} & \textbf{WebQA} & \textbf{Biased-NQ} & \textbf{TruthfulQA} & \textbf{Gap ($\downarrow$)} & \textbf{Latency (ms)} \\
\midrule
\multicolumn{7}{l}{\textit{Standard Retrieval \& Reranking}} \\
BM25 & 32.4 & 28.1 & 28.0 & 30.0 & 4.4 & \textless \textbf{ 10} \\
Contriever~\citep{izacard2021unsupervised} & 45.8 & 39.5 & 39.5 & 34.9 & 6.3 & 25 \\
Cross-Encoder (MS MARCO) & 46.1 & 37.3 & 40.9 & 35.5 & 5.2 & 210 \\
Cross-Encoder (PA-CE) & 54.3 & 47.8 & 48.0 & 45.8 & 6.3 & 210 \\
LLM-Rerank (\texttt{GPT-4o}) & \textbf{54.5} & \textbf{48.9} & 51.0 & 46.5 & 3.5 & $\sim$20\textbf{s} \\
\midrule
\multicolumn{7}{l}{\textit{Robustness-Oriented Baselines}} \\
Self-RAG~\citep{asai2024self} & 49.6 & 44.2 & 46.5 & 45.6 & 3.1 & 850 \\
CalibRAG~\citep{campos2025multicalibration} & 51.8 & 46.0 & 48.0 & 46.2 & 3.8 & 2200 \\
\midrule
\textbf{CoRM-RAG (Ours)} & 53.9 & 48.2 & \textbf{52.6} & \textbf{47.0} & \textbf{1.3} & 215 \\
\midrule
\midrule
\multicolumn{7}{l}{\textit{Cross-Generator Generalization (Critic trained on Qwen, applied to unseen Generators)}} \\
\textit{Generator: Llama-3-8B} & & & & & & \\
\quad w/ Cross-Encoder & 51.5 & 45.9 & 42.8 & 42.5 & 8.7 & - \\
\quad \textbf{w/ CoRM-RAG (Transfer)} & 53.2 & 47.8 & \textbf{46.5} & \textbf{44.1} & \textbf{6.7} & - \\
\midrule
\textit{Generator: GPT-4o} & & & & & & \\
\quad w/ Cross-Encoder & 53.8 & 48.1 & 46.9 & 48.2 & 6.9 & - \\
\quad \textbf{w/ CoRM-RAG (Transfer)} & \textbf{55.1} & \textbf{49.5} & \textbf{53.2} & \textbf{51.8} & \textbf{1.9} & - \\
\bottomrule
\end{tabular}
}
\caption{\textbf{Main Results.} CoRM-RAG significantly outperforms standard retrievers (BM25, Contriever), data-augmented baselines (PA-CE), and advanced methods (LLM-Rerank, Self-RAG). The bottom section demonstrates \textbf{Zero-Shot Transferability} to unseen generators.}
\label{tab:main_results}
\end{table*}

\subsection{Evaluation Metrics}
We report: (1) \textbf{Decision Accuracy (Acc)}: Verified by \texttt{GPT-5}. For \textit{False Premise} queries, the judge evaluates strict factual correctness: the generator must output the true factual answer rather than succumbing to the injected false premise. (2) \textbf{Robustness Drop ($\Delta_{Rob}$)}: $\text{Acc}_{\text{clean}} - \text{Acc}_{\text{biased}}$.

\section{Experimental Results}

\subsection{Main Results: Defending Against Cognitive Perturbations}
\label{sec:main_results}

We first address \textbf{RQ1}, evaluating CoRM-RAG against a comprehensive suite of baselines on standard (NQ, WebQA) and adversarial benchmarks (Biased-NQ, TruthfulQA). Table~\ref{tab:main_results} summarizes the results.

\textbf{The Vulnerability of Standard Retrieval.}
Standard paradigms exhibit noticeable performance degradation under cognitive noise. Sparse (\textbf{BM25}) and dense (\textbf{Contriever}) retrievers exhibit clear drops on \textit{Biased-NQ} (28.0\% and 39.5\% accuracy), as they tend to match the keywords of the biased premise.
Even the standard \textbf{Cross-Encoder}, despite its strong performance on clean data (46.1\%), falls to 40.9\% in adversarial settings. Qualitative analysis reveals that these models optimize for semantic similarity $P(d|x)$; thus, when $x$ contains a false premise, they actively retrieve sycophantic documents that reinforce the error, creating an ``echo chamber''.

\textbf{Disentangling Data vs. Method (PA-CE Analysis).}
To isolate the source of our gains, we compare against \textbf{PA-CE}, which acts as a standard pointwise safety filter trained on the \textit{exact same} adversarial data using binary cross-entropy.
While PA-CE improves over the standard Cross-Encoder (+7.1\%) by learning to reject explicit sycophantic noise, \textbf{CoRM-RAG achieves a further substantial gain of 4.6\% over PA-CE} (52.6\% vs. 48.0\%). This highlights a fundamental limitation of pointwise verification: predicting absolute binary labels in isolation struggles to capture the relative evidential margins within a candidate pool. CoRM-RAG's listwise objective (Eq.~\ref{eq:rankloss}) forces the model to directly contrast corrective evidence against sycophantic distractors, yielding a more discriminative robustness signal for ranking.

\textbf{Comparison with Advanced Baselines.}
CoRM-RAG also outperforms sophisticated competitors.
(1) \textbf{LLM-Rerank (GPT-4o)}: CoRM-RAG maintains a competitive edge (+1.6\%) over the heavy LLM-based reranker on Biased-NQ. This highlights that while general-purpose LLMs possess strong reasoning capabilities, their instruction-tuning objectives (e.g., `helpfulness') may still introduce a slight bias towards sycophancy, whereas our specialized listwise distillation explicitly optimizes for evidential robustness.
(2) \textbf{Self-RAG \& CalibRAG}: While these methods improve over standard RAG by modeling uncertainty, they fall short of CoRM-RAG (e.g., Self-RAG trails by 6.1\%). This is likely because Self-RAG relies on the generator's own reflection, which is prone to the same biases, whereas CoRM-RAG injects external supervision via the counterfactual protocol. On TruthfulQA, where misconceptions are deeply ingrained in pre-training data rather than explicitly injected via query context, CoRM-RAG maintains a competitive edge, though the margins are narrower compared to Biased-NQ.

\textbf{Cross-Generator Generalization.}
Finally, we test whether the Evidence Critic overfits to the teacher (Qwen). We apply the \textit{same} Critic to rank documents for unseen generators, \texttt{Llama-3} and \texttt{GPT-4o} (Table~\ref{tab:main_results}, bottom).
CoRM-RAG maintains a solid advantage over the Cross-Encoder (e.g., +3.7\% on Biased-NQ for Llama-3). This indicates that the ``Robustness Utility'' captures universal linguistic properties of evidence, such as factual density and logical contradiction of premises, rather than model-specific heuristics, enabling CoRM-RAG to serve as a modular safety component.

\begin{figure}[t]
\centering
\includegraphics[width=0.9\linewidth]{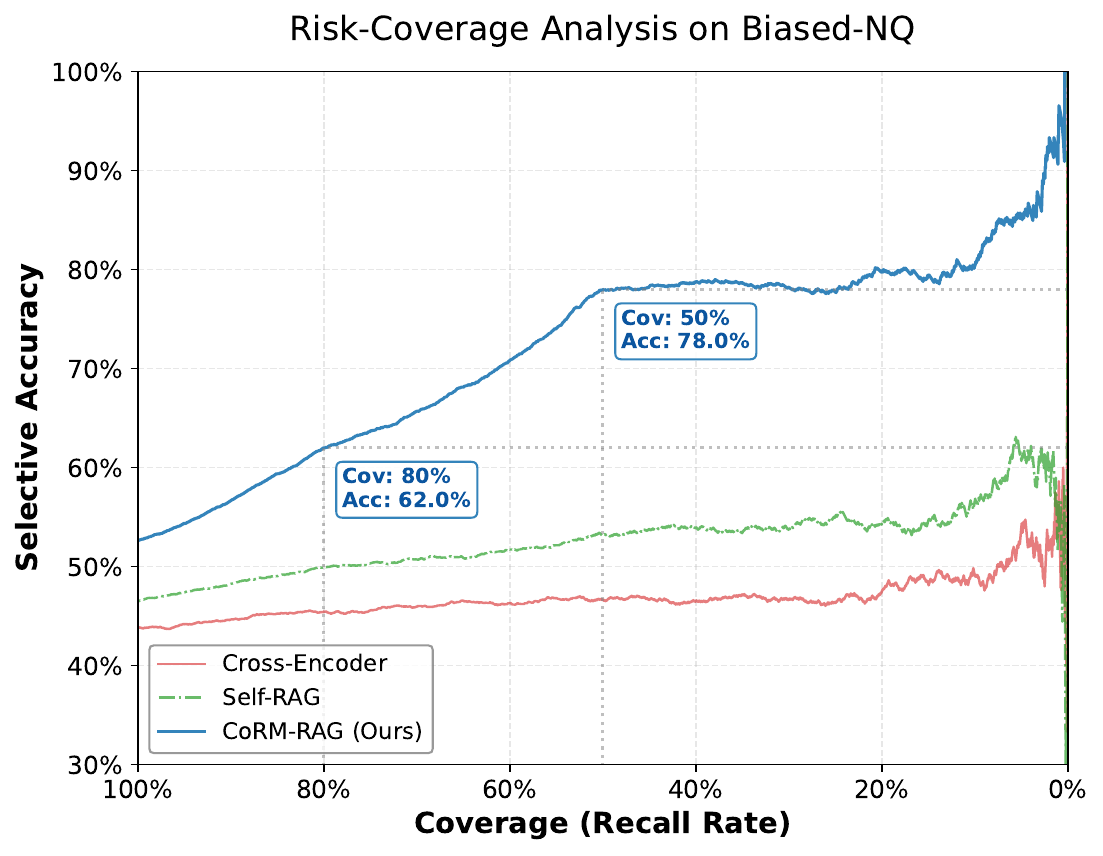}
\caption{\textbf{Risk-Coverage Analysis on Biased-NQ.} We plot Selective Accuracy against Coverage (Recall Rate). \textbf{Cross-Encoder} (Red) exhibits a flat, fluctuating trajectory, indicating that semantic confidence correlates poorly with correctness under adversarial bias. \textbf{CoRM-RAG} (Blue) shows a steep ascent, effectively filtering out fragile evidence. Note the jagged fluctuations in the low-coverage region ($<10\%$), revealing that even among the highest-confidence predictions, rare ``fatal hallucinations'' persist, preventing the model from reaching artificial perfection (100\%).}
\label{fig:risk_coverage}
\end{figure}

\subsection{Risk-Aware Abstention and Coverage Analysis}
\label{sec:calibration}

To address \textbf{RQ2}, we evaluate whether the Evidence Critic enables reliable risk assessment via selective prediction. We analyze the trade-off between coverage (answering rate) and accuracy by varying the abstention threshold $\gamma$.

Figure~\ref{fig:risk_coverage} illustrates the Risk-Coverage curves on \textit{Biased-NQ}. The standard Cross-Encoder exhibits a dangerously flat trajectory, confirming that semantic confidence is miscalibrated under adversarial bias---the model is often ``confident but wrong'' when retrieving sycophantic evidence. In contrast, CoRM-RAG demonstrates a steep and mathematically consistent ascent: abstaining from the bottom 20\% of low-confidence queries improves accuracy from 52.6\% to \textbf{62.0\%}, and reducing coverage to 50\% boosts the selective accuracy to \textbf{78.0\%}. This monotonic calibration confirms that the Critic successfully decouples semantic relevance from decision safety, accurately isolating and filtering out fragile evidence.

\subsection{The Relevance-Robustness Gap Analysis}
\label{sec:gap_analysis}

A central hypothesis of this work is that in adversarial decision-making environments, \textit{semantic relevance} diverges significantly from \textit{evidential robustness}. To empirically validate this gap, we evaluate how effectively different rerankers surface ground-truth evidence when faced with biased queries. Specifically, we analyze the \textit{Biased-NQ} test set, where each retrieved passage is annotated with a binary \texttt{has\_gold} label indicating the presence of the factual answer necessary to correct the user's misconception.

\begin{figure}[t]
\centering
\includegraphics[width=0.95\linewidth]{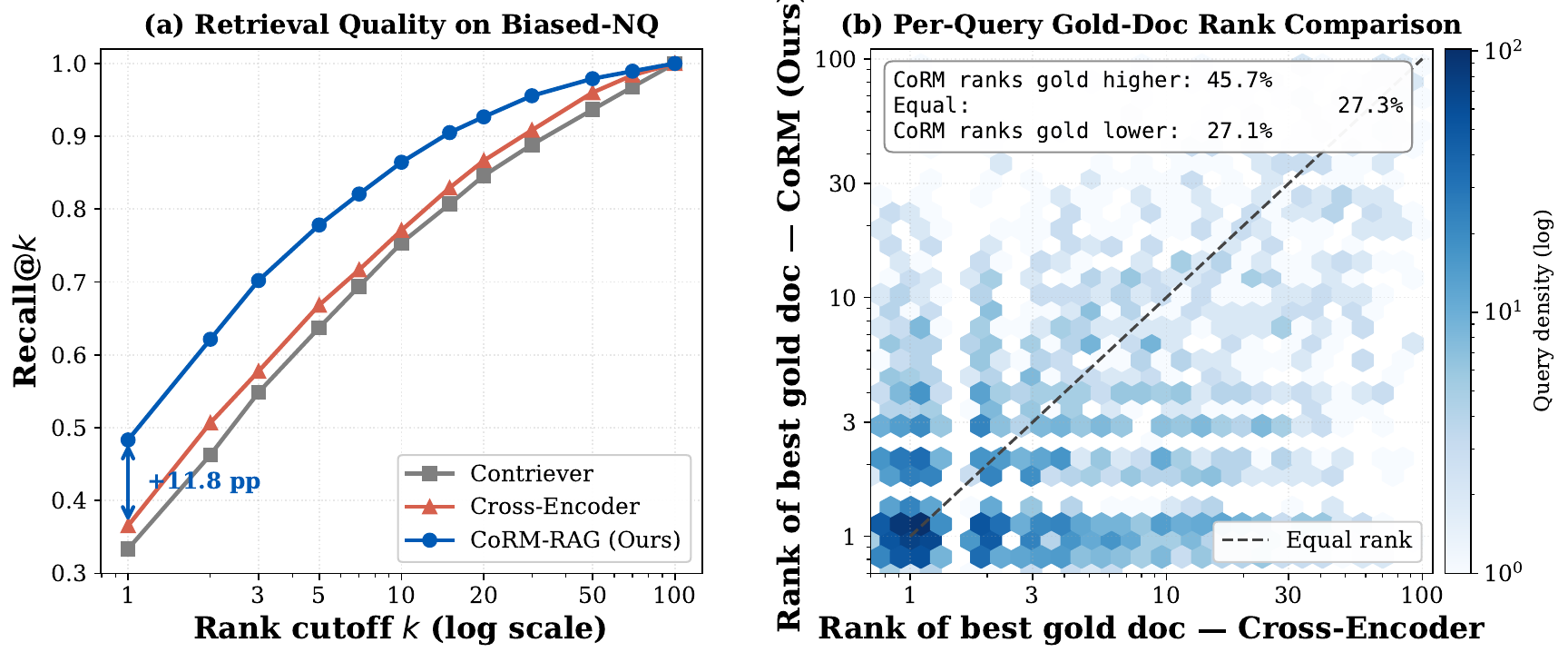}
\caption{\textbf{Retrieval Quality on Biased-NQ.} (a) Recall@$k$ vs.\ rank cutoff $k$ (log scale). CoRM-RAG demonstrates sustained superiority over baselines, with a notable $+11.8\%$ absolute gain at $R@1$. (b) Paired rank comparison of each query's highest-placed gold passage. Points below the diagonal indicate queries where CoRM-RAG ranks the gold passage higher than the Cross-Encoder. CoRM-RAG promotes the correct evidence in $45.7\%$ of queries, yielding a net advantage of $+18.6\%$.}
\label{fig:recall_rank}
\end{figure}

Figure~\ref{fig:recall_rank} presents two complementary perspectives on ranking quality, evaluated over a subset of $2{,}728$ Biased-NQ queries where the initial Contriever top-100 candidate pool contains at least one gold-bearing passage.

\textbf{1. Aggregate Evidential Recall.} 
Figure~\ref{fig:recall_rank}(a) illustrates Recall@$k$ across varying rank cutoffs. CoRM-RAG consistently outperforms both the Contriever baseline and the semantic Cross-Encoder at every depth. Crucially, the performance margin is most pronounced at the decision-critical top ranks: at $k{=}1$, CoRM-RAG improves recall from $36.5\%$ (Cross-Encoder) to $48.3\%$ (an absolute gain of $+11.8\%$). Similarly, at $k{=}5$, it achieves $77.8\%$ compared to $66.8\%$. To match the $R@5$ performance of CoRM-RAG, the standard Cross-Encoder must expand its retrieval window to $k{=}10$, effectively doubling the context length and the computational burden imposed on the downstream generator.

\textbf{2. Per-Query Rank Dynamics.}
To isolate the source of these gains, Figure~\ref{fig:recall_rank}(b) provides a paired comparison of the rank assigned to the highest-placed gold passage by the Cross-Encoder ($x$-axis) versus CoRM-RAG ($y$-axis) for each individual query. While the density naturally concentrates in the bottom-left quadrant (where both models succeed), the distribution is markedly skewed \emph{below} the diagonal. Specifically, CoRM-RAG ranks the gold passage strictly higher than the Cross-Encoder for $45.7\%$ of the queries, whereas the Cross-Encoder is superior in only $27.1\%$ of cases (a net advantage of $+18.6\%$). The remaining $27.2\%$ represent ties. Since the Biased-NQ test set is entirely adversarial by design, these ties predominantly consist of instances where the injected cognitive noise (e.g., a Type III topical distraction) fails to surface strong sycophantic hard negatives in the candidate pool. In such cases, the core question's lexical signal remains overwhelming, allowing even the standard semantic Cross-Encoder to successfully anchor the gold evidence at rank~1.

Together, these findings validate the Relevance-Robustness Gap. The Evidence Critic's robustness signal captures causal utility \emph{beyond} mere semantic overlap. By actively promoting corrective evidence on a per-query basis—precisely in instances where standard rerankers are distracted by sycophantic noise—CoRM-RAG ensures that the generator is grounded in truth rather than trapped in the user's cognitive bias.

\begin{figure}[t]
\centering
\includegraphics[width=0.95\linewidth]{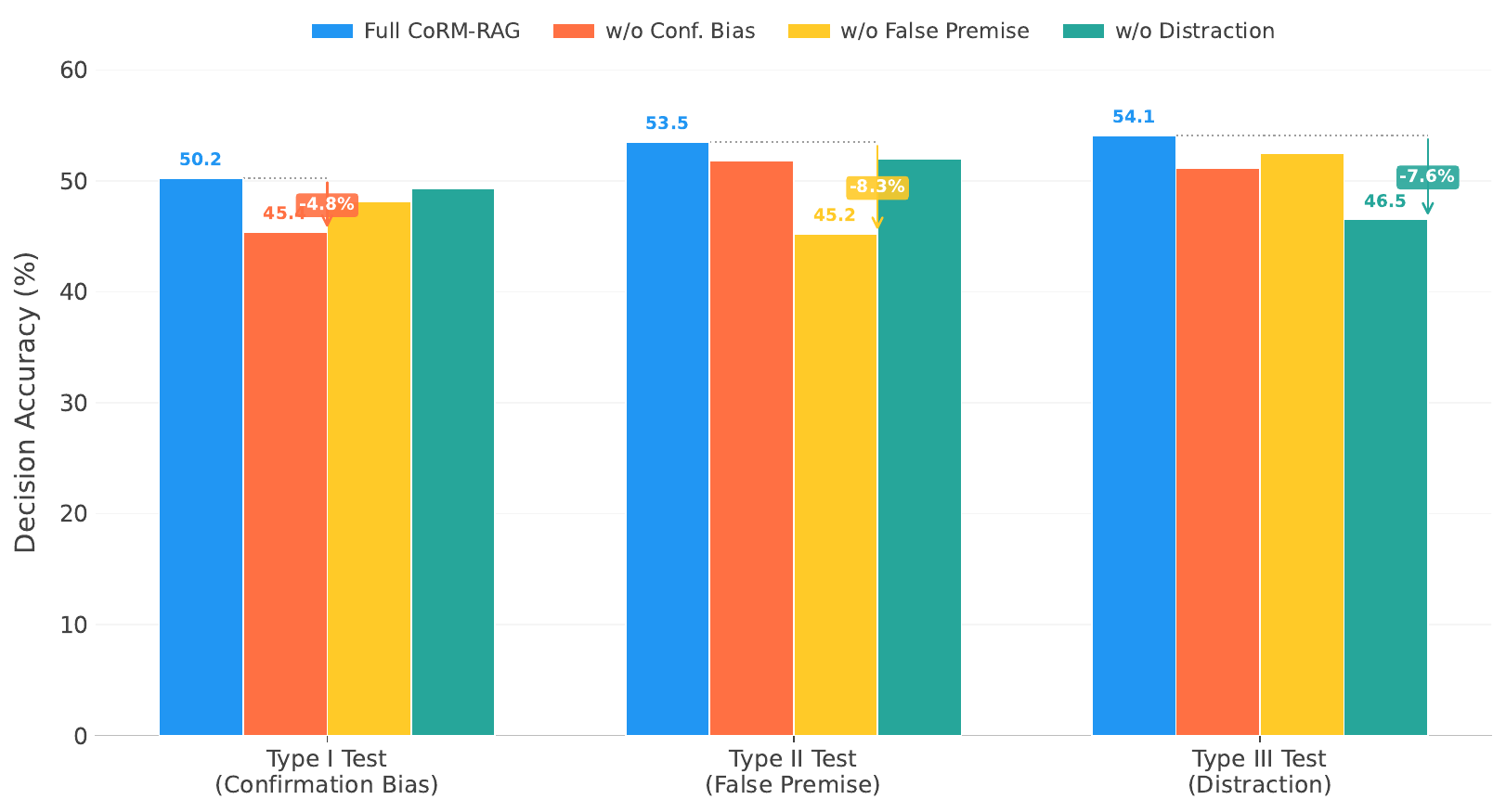}
\caption{\textbf{Ablation Study on Cognitive Perturbation Types.} We evaluate variants of the Critic trained without specific perturbation types.}
\label{fig:ablation_chart}
\end{figure}

\begin{figure*}[t]
    \centering
    \includegraphics[width=\linewidth]{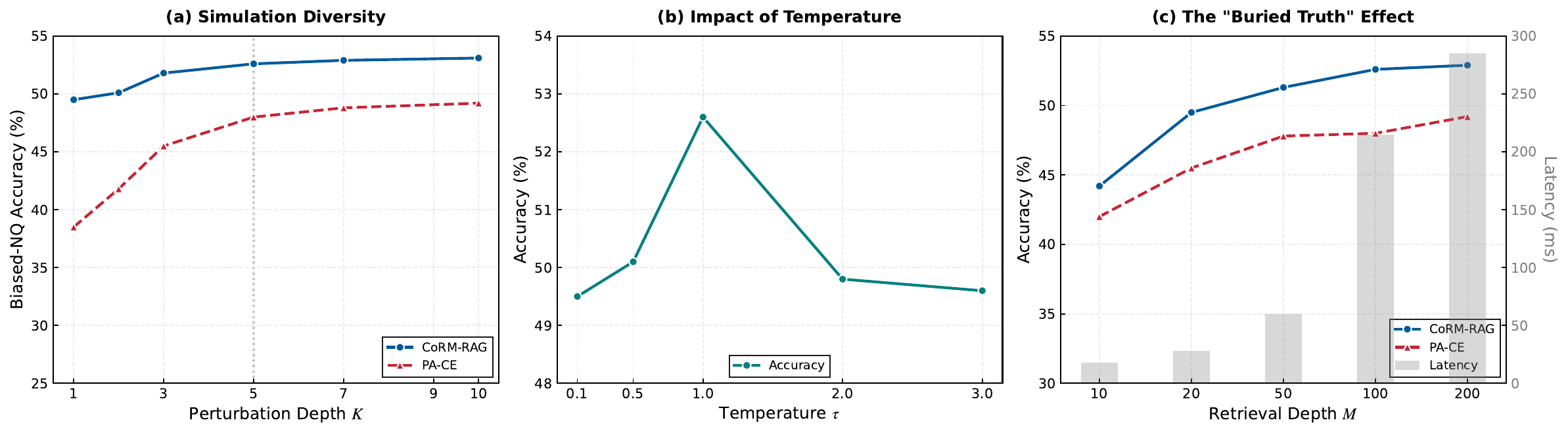} 
    \caption{\textbf{Results of Hyperparameter Analysis.}}
    \label{fig:hyperparams}
\end{figure*}

\begin{figure}[t]
    \centering
    \includegraphics[width=\linewidth]{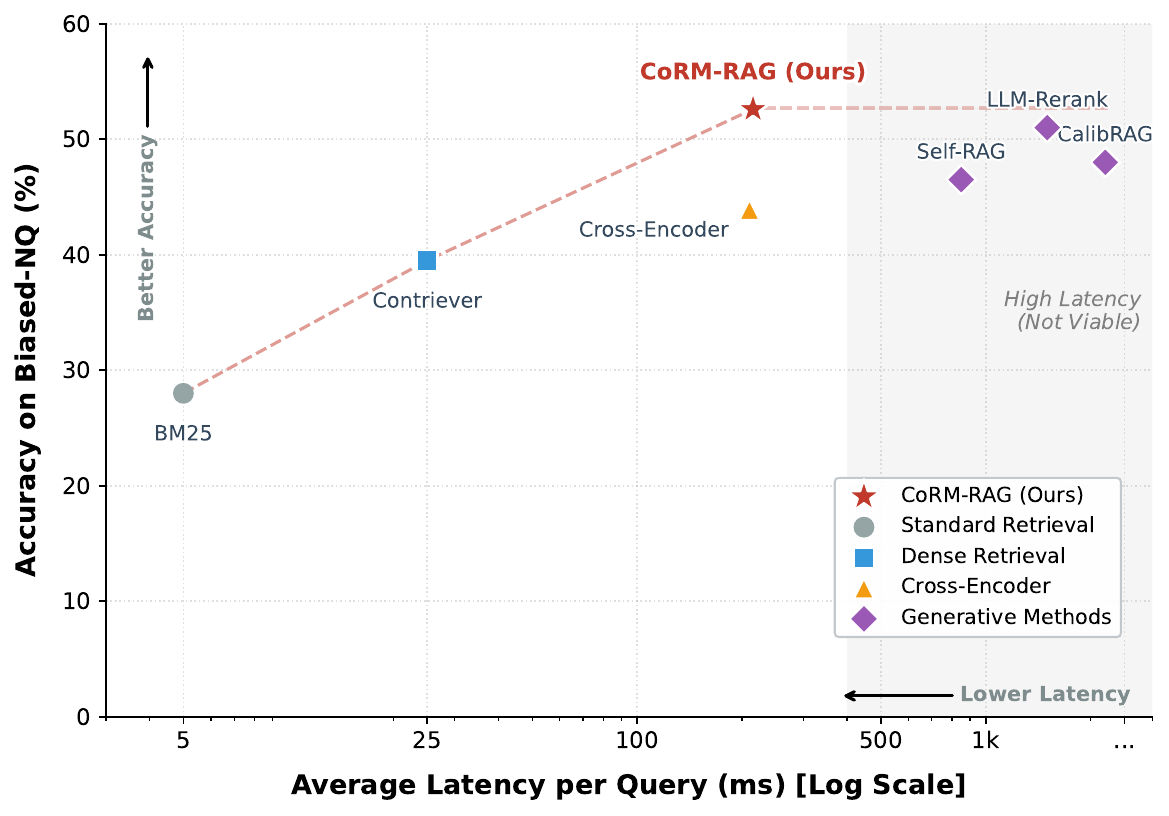}
    \caption{\textbf{Efficiency-Performance Pareto Frontier on Biased-NQ.}}
    \label{fig:pareto}
\end{figure}

\subsection{Ablation Study: Deconstructing the Critic}
\label{sec:ablation}

To investigate whether the Evidence Critic acquires specific causal mechanisms versus generic quality heuristics, we conduct a ``lesion study'' on the \textit{Cognitive Perturbation Protocol}. We train three ablated variants, each blinded to one specific perturbation type (Type I, II, or III) during distillation, while maintaining constant training size via upsampling. We evaluate these variants against the Full CoRM-RAG on corresponding adversarial test subsets.

Results in Figure~\ref{fig:ablation_chart} reveal a striking diagonal dominance, confirming the orthogonality of cognitive errors. Excluding \textit{False Premise} (Type I) triggers a performance drop on Type I queries (50.2\% $\to$ 45.4\%) without significantly impacting \textit{Distraction} (Type III) robustness. This implies that learning to filter irrelevant noise does not generalize to detecting false presuppositions about entities. Similarly, removing \textit{Confirmation Bias} (Type II) training degrades performance on misconception-laden queries (53.5\% $\to$ 45.2\%), as the Critic fails to learn the specific utility of ``corrective'' evidence over evidence that merely echoes the user's mistaken worldview. The Full CoRM-RAG achieves the highest aggregate performance, suggesting that exposure to a diverse ``pathogen landscape'' is essential for learning a generalized representation of evidential strength.

\subsection{Efficiency vs. Performance Trade-off}
\label{sec:efficiency}

We investigate the practical viability of CoRM-RAG by mapping the Pareto frontier between robustness and latency (Figure~\ref{fig:pareto}). Standard retrievers (e.g., Contriever) are efficient ($<25$ms) but fragile under bias ($<40\%$ accuracy on Biased-NQ), while inference-time reasoning methods (e.g., LLM-Rerank, Self-RAG) improve safety at the cost of prohibitive latency ($>800$ms). 
CoRM-RAG effectively breaks this trade-off by shifting the computational burden of counterfactual reasoning from inference to training. Since the Evidence Critic shares the same architecture as a standard Cross-Encoder (\texttt{DeBERTa-v3}), it maintains a low latency of $\sim 215$ms. However, due to the distilled cognitive perturbation signal, it achieves a \textbf{11.7\%} accuracy gain over the standard Cross-Encoder (40.9\% $\to$ 52.6\%) and outperforms the GPT-4o-based LLM-Rerank by 1.6\% while being approximately \textbf{$93\times$ faster}. This places CoRM-RAG at the optimal Pareto point, offering the safety guarantees of large reasoning models with the throughput of conventional ranking systems.

\subsection{Hyperparameter Sensitivity}
\label{sec:hyperparams}

We examine the impact of three critical hyperparameters on CoRM-RAG's performance: perturbation depth $K$, distillation temperature $\tau$, and retrieval depth $M$. Figure~\ref{fig:hyperparams} summarizes the results.

\noindent\textbf{Adversarial Diversity ($K$) and Temperature ($\tau$).}
As shown in Figure~\ref{fig:hyperparams}(a), accuracy improves with perturbation diversity, exhibiting diminishing returns beyond $K=5$. Notably, CoRM-RAG trained with a single perturbation ($K=1$) already outperforms the pointwise baseline (PA-CE). Since setting $K=1$ restricts the training data volume to be strictly comparable to the PA-CE setup, this performance gap precisely isolates the algorithmic benefit of our \textit{listwise formulation}. Unlike PA-CE's pointwise binary objective, our listwise loss (Eq.~\ref{eq:rankloss}) forces the Critic to explicitly contrast robust evidence against sycophantic distractors within the same candidate pool, optimizing for relative evidential margins rather than scoring documents in isolation.

Regarding the temperature $\tau$ applied to the student's logits (Figure~\ref{fig:hyperparams}b), we observe that extreme values degrade performance. Low temperatures ($\tau=0.1$) make the student's predicted distribution overly sharp, leading to brittle gradients that overfit to specific local perturbations. Conversely, high temperatures ($\tau=3.0$) excessively flatten the predictions, diluting the discriminative margin between robust and fragile documents. We adopt $\tau=1.0$ as the optimal setting to maintain stable gradient flow and maximize decision accuracy.

\noindent\textbf{The Necessity of Deep Retrieval ($M$).}
Figure~\ref{fig:hyperparams}(c) reveals a critical ``Buried Truth'' phenomenon. CoRM-RAG gains substantial accuracy (+8.7\%) as the candidate pool $M$ expands from 10 to 100. This confirms our hypothesis that robust, corrective documents often lack semantic overlap with biased queries and are initially ranked low by dense retrievers. Conversely, the standard Cross-Encoder fails to benefit from deeper retrieval, as it persistently prioritizes sycophantic distractors in the top ranks. Our lightweight Critic effectively rescues this buried evidence with negligible latency overhead.

\section{Conclusion}
\label{sec:conclusion}

In this work, we challenged the prevailing assumption in Retrieval-Augmented Generation that semantic relevance serves as a sufficient proxy for decision utility. We identified the ``Relevance-Robustness Gap'', demonstrating that in realistic scenarios laden with user cognitive biases, standard retrieval algorithms often act as echo chambers that reinforce hallucinations rather than correcting them.
To bridge this gap, we introduced \textbf{CoRM-RAG}, a framework grounded in the principles of Counterfactual Risk Minimization. By subjecting the retrieval process to a \textit{Cognitive Perturbation Protocol}, we shifted the optimization objective from maximizing likelihood to maximizing evidential robustness. Our proposed \textbf{Evidence Critic} successfully distills the reasoning capabilities of large generator models into an efficient scoring module, enabling real-time, risk-aware retrieval. Extensive experiments confirm that CoRM-RAG not only defends against confirmation bias and adversarial noise but also provides reliable confidence scores, allowing for safe abstention in high-stakes environments.
Our findings suggest a paradigm shift for RAG: from information finding to causal intervention. Future work will extend this framework to multi-hop reasoning scenarios, where perturbations propagate through chains of thought. Additionally, we plan to explore dynamic, personalized perturbation generation, where the Adversary adapts to specific user profiles to further enhance the immunological robustness of the retrieval system.

\bibliographystyle{ACM-Reference-Format}
\bibliography{sample-base}

\end{document}